\documentclass[10pt,conference]{IEEEtran}

\usepackage{multirow,makecell}
\usepackage{tcolorbox}
\usepackage{color,xcolor}
\usepackage{listings,amsfonts}
\usepackage{caption}
\usepackage{subcaption}
\usepackage{threeparttable}
\usepackage{bbding}
\usepackage{graphicx}
\usepackage{booktabs} 
\usepackage{longtable}
\usepackage{pifont}
\usepackage{amsmath}
\usepackage[linesnumbered,ruled,vlined]{algorithm2e}
\usepackage{amssymb}
\usepackage{enumitem}

\definecolor{customcite}{HTML}{b83b5e}
\definecolor{customlink}{HTML}{07689f}
\definecolor{customurl}{HTML}{ff8264}
\AtEndPreamble{
\usepackage{hyperref}

    \hypersetup{
      colorlinks = true,
      linkcolor = customlink,
      anchorcolor = purple,
      citecolor = customlink,
      filecolor = purple,
      urlcolor = customlink
    }
}

\newcommand{\tool}[1]{\textsc{TensorGuard}}

\newcommand{\find}[1]{
\begin{tcolorbox}[leftrule=0.5mm,toprule=0mm,bottomrule=0mm,left=0.7pt,right=0.7pt,top=0.2pt,bottom=0.2pt]
\em #1
\end{tcolorbox}
}

\title{Gradient-Based Model Fingerprinting for LLM Similarity Detection and Family Classification}

\author{
\IEEEauthorblockN{Zehao Wu$^{1,3}$, Yanjie Zhao$^{1,3}$, and Haoyu Wang$^{2,3}$}
\IEEEauthorblockA{
Huazhong University of Science and Technology, Wuhan, China\\
wuzehao195@hust.edu.cn, yanjie\_zhao@hust.edu.cn, haoyuwang@hust.edu.cn}
}

\begin{document}

\maketitle

\footnotetext[1]{Zehao Wu and Yanjie Zhao contributed equally to this work.}
\footnotetext[2]{Haoyu Wang is the corresponding author (haoyuwang@hust.edu.cn).}
\footnotetext[3]{The full name of the authors' affiliation is Hubei Key Laboratory of Distributed System Security, Hubei Engineering Research Center on Big Data Security, School of Cyber Science and Engineering, Huazhong University of Science and Technology.}

\maketitle
\begin{abstract}
As Large Language Models (LLMs) become integral software components in modern applications, unauthorized model derivations through fine-tuning, merging, and redistribution have emerged as critical software engineering challenges. Unlike traditional software where clone detection and license compliance are well-established, the LLM ecosystem lacks effective mechanisms to detect model lineage and enforce licensing agreements. This gap is particularly problematic when open-source model creators, such as Meta's LLaMA, require derivative works to maintain naming conventions for attribution, yet no technical means exist to verify compliance.

To fill this gap, treating LLMs as software artifacts requiring provenance tracking, we present \tool{}, a gradient-based fingerprinting framework for LLM similarity detection and family classification. Our approach extracts model-intrinsic behavioral signatures by analyzing gradient responses to random input perturbations across tensor layers, operating independently of training data, watermarks, or specific model formats. \tool{} supports the widely-adopted \texttt{safetensors} format and constructs high-dimensional fingerprints through statistical analysis of gradient features. These fingerprints enable two complementary capabilities: direct pairwise similarity assessment between arbitrary models through distance computation, and systematic family classification of unknown models via the K-Means clustering algorithm with domain-informed centroid initialization using known base models.
Experimental evaluation on 58 models comprising 8 base models and 50 derivatives across five model families (Llama, Qwen, Gemma, Phi, Mistral) demonstrates 94\% classification accuracy under our centroid-initialized K-Means clustering. Our work establishes a new paradigm for model similarity detection, bridging traditional software engineering practices with modern LLM distribution and compliance challenges.

\end{abstract}

\section{Introduction}
\label{sec:introduction}

The proliferation of Large Language Models (LLMs) has fundamentally transformed how we conceptualize and deploy AI-powered software systems. With over one million model repositories on platforms like Hugging Face~\cite{huggingface_models}, LLMs have evolved from research artifacts into critical software components powering applications from code generation to intelligent assistants. However, this transformation has exposed a significant gap in software engineering practices: while traditional software enjoys mature ecosystems for clone detection~\cite{shobha2021code}, license compliance~\cite{scacchi2012understanding}, and intellectual property protection, \textbf{the LLM domain lacks equivalent mechanisms for model lineage tracking, derivative work detection, and architectural family classification}.

This challenge is exemplified by licensing requirements from major open-source LLM providers. Meta's LLaMA 3 license explicitly requires derivative models to retain ``Llama 3'' in their naming~\cite{llama_license}, while other providers impose similar attribution requirements~\cite{gemma3_terms}. Yet unlike traditional software where static analysis tools can detect code reuse and licensing violations~\cite{pistoia2007survey}, \textbf{the model ecosystem operates without technical means to verify compliance or classify models into their originating families}. This regulatory vacuum is concerning given that as of July 2024, only 37\% of publicly released models provided license information~\cite{gretel_slm_license}, suggesting widespread non-compliance with attribution requirements.

The technical challenges of LLM similarity detection and family classification differ fundamentally from traditional software analysis. While conventional approaches analyze syntactic patterns, control flow graphs, or semantic representations of source code~\cite{pistoia2007survey}, LLM analysis must capture behavioral and architectural characteristics encoded in high-dimensional parameter spaces through model fingerprinting techniques. Models undergo complex transformations through fine-tuning, parameter merging, and quantization that alter their internal representations while potentially preserving core architectural DNA from their base models, making \textbf{both similarity detection and family classification inherently dependent on robust fingerprinting methodologies}.

Existing approaches to model fingerprinting fall into three categories, each with significant limitations for practical applications. \textbf{(1) Watermarking-based methods}~\cite{xu2024instructional,russinovich2024hey} require pre-deployment modifications that may degrade model performance and can only be utilized by original model owners, rendering them unusable for third-party auditors. \textbf{(2) Output-based fingerprinting}~\cite{yang2024fingerprint,mcgovern2024your} relies on behavioral analysis through prompting but struggles with fine-tuned models and suffers from generation randomness that makes it difficult to establish stable fingerprints. \textbf{(3) Internal feature analysis methods}~\cite{zhang2024reef,zeng2024huref} show promise for similarity detection, while approaches like MoTHer~\cite{horwitz2024origin} attempt family classification through heritage recovery. However, these methods exhibit critical limitations: poor compatibility with the dominant \texttt{safetensors} format~\cite{huggingface_models_safetensors}, restricted support for specific model families (MoTHer supports only LLaMA2 and Stable Diffusion), lack of open-source availability, and assumptions about tensor layouts that may not generalize across diverse model architectures encountered in practice.

To bridge this gap, we propose \tool{}, a gradient-based fingerprinting framework specifically designed for LLM similarity detection and family classification. Our approach treats LLMs as software artifacts requiring provenance tracking, drawing inspiration from traditional clone detection while addressing the unique challenges of neural model analysis. \tool{} extracts model fingerprints by analyzing gradient responses to controlled input perturbations across tensor layers, capturing intrinsic behavioral characteristics that persist through common model modification techniques. Unlike existing approaches, our method operates independently of training data, embedded watermarks, and provides native support for the \texttt{safetensors} format, making it suitable for third-party auditing and compliance verification scenarios common in software engineering workflows.

Our key contributions are shown as follows:

\begin{itemize}[leftmargin=*]
    \item \textbf{Novel gradient-based model fingerprinting}: We introduce a perturbation-driven approach that captures model-intrinsic behavioral signatures through structured gradient analysis, creating unique fingerprints that enable both direct pairwise similarity assessment between arbitrary models and systematic family classification, contributing to the challenge of model provenance tracking in modern LLM systems.
    
    \item \textbf{Dual-capability detection framework}: We implement \tool{} with native support for \texttt{safetensors} format and an extensible architecture that supports two complementary applications: distance-based similarity measurement for any model pair, and centroid-initialized clustering for classifying unknown models into established architectural families using known base models as reference points.
    
    \item \textbf{Comprehensive empirical validation}: We evaluate our approach on 58 models comprising 8 base models and 50 derivatives spanning five major model families (LLaMA, Qwen, Gemma, Phi, and Mistral). Our centroid-initialized K-Means clustering achieves 94\% accuracy in family classification, while our distance-based similarity measurement provides effective benchmarks for gradient-based model comparison across arbitrary model pairs.
    
\end{itemize}

This work establishes a solid foundation for treating model similarity detection as a core software engineering practice, enabling the systematic development of automated compliance tools, robust license verification systems, and comprehensive intellectual property protection mechanisms that are essential for the secure and mature deployment of LLM-based software systems. 

\section{Background}
\label{sec:background}

\subsection{\texttt{Safetensors} Format}
\label{sec:safetensors}

As the deployment of LLMs becomes increasingly prevalent, ensuring the security and efficiency of model serialization formats has emerged as a paramount concern. Conventional model formats, including \texttt{.bin}, \texttt{.pt}, and \texttt{.pth} files, typically employ Python's \texttt{pickle} module for serialization, thereby introducing severe security vulnerabilities due to the potential for arbitrary code execution during the deserialization process~\cite{huang2022pain}. To address these critical security concerns, the Hugging Face community developed the \texttt{safetensors} format, which enhances security while preserving zero-copy and lazy-loading capabilities~\cite{safetensors}.

The \texttt{safetensors} format employs a well-defined binary structure: an initial 8-byte header specifies the length of a UTF-8 encoded JSON metadata section, which contains essential tensor information including names, shapes, data types, and byte offsets. The subsequent portion of the file stores the actual tensor data in binary format~\cite{safetensors}. This architectural design achieves a clear separation between data and executable code, thereby eliminating the code injection vulnerabilities inherent in \texttt{pickle}-based serialization schemes.
Relative to alternative formats, \texttt{safetensors} demonstrates substantial advantages across multiple dimensions: security, memory efficiency, and compatibility with large-scale model deployment pipelines. Notable features include support for lazy loading mechanisms, fine-grained layout control for grouped tensors, and hardware-efficient formats such as BF16 and FP8. Comparative analyses of existing model formats indicate that \texttt{safetensors} uniquely achieves the dual objectives of security and performance without compromising flexibility~\cite{safetensors}.

Nevertheless, despite its enhanced security architecture, the \texttt{safetensors} format remains vulnerable to certain format-level attacks, including denial-of-service exploits through overlapping tensor offsets or malformed tensor specifications. Recommended mitigation strategies encompass rigorous offset validation and comprehensive range checking during model loading procedures. As of 2025, the format has achieved widespread adoption, with over 700,000 models on the Hugging Face platform utilizing \texttt{safetensors} for distribution~\cite{huggingface_models_safetensors}, establishing it as the de facto standard for secure LLM storage and distribution.

\subsection{Fine-tuning Techniques for LLMs}

Pre-trained LLMs are typically developed through training on extensive corpora using general-purpose objectives. While these models exhibit robust generalization capabilities, their outputs frequently fail to satisfy task-specific requirements when deployed directly in specialized domains. For example, when queried for legal counsel, an LLM may generate hallucinated or fabricated regulations due to insufficient exposure to domain-specific legal datasets during pre-training. To mitigate such limitations, fine-tuning methodologies have been developed to adapt pre-trained models for downstream applications, thereby enhancing both task performance and alignment with user expectations~\cite{dodge2020fine}.
Fine-tuning techniques can be systematically categorized into four principal paradigms based on their underlying mechanisms: full-parameter fine-tuning, parameter-efficient fine-tuning~\cite{xu2023parameter}, prompt-based tuning~\cite{lester2021power}, and reinforcement learning-based fine-tuning~\cite{ziegler2019fine}. 

\textbf{Full-Parameter Fine-Tuning (FFT)} involves updating all parameters of a pre-trained model and represents the foundational fine-tuning methodology. The transfer learning framework introduced with BERT~\cite{devlin2019bert} exemplifies this technique. While this method achieves superior accuracy, it incurs substantial computational and storage costs, particularly when applied to large-scale models containing billions of parameters.
\textbf{Parameter-Efficient Fine-Tuning (PEFT)} seeks to minimize resource requirements by adjusting only a limited subset of parameters or incorporating lightweight architectural components. Representative techniques include Adapter Tuning~\cite{houlsby2019parameter}, Prefix-Tuning~\cite{li2021prefix}, and Low-Rank Adaptation (LoRA)~\cite{hu2022lora}. Among these approaches, LoRA has garnered considerable attention due to its theoretical foundation and practical effectiveness. LoRA exploits the observation that pre-trained models exhibit low intrinsic dimensionality~\cite{aghajanyan2020intrinsic} and adapts the model by injecting low-rank decomposition matrices into frozen pre-trained weights. Analogous to LoRA, other parameter-efficient methods employ the strategy of freezing the majority of model parameters while fine-tuning a minimal set of trainable components, thereby substantially reducing training overhead~\cite{xu2023parameter}.
\textbf{Prompt-Based Tuning} involves designing task-specific prompts to guide model behavior without modifying the underlying parameters. A closely related approach, instruction tuning, utilizes comprehensive datasets of human-authored or synthetically generated instructions to enhance the model's generalization capacity and zero-shot performance across diverse tasks~\cite{liu2023visual}.
\textbf{Reinforcement Learning-Based Fine-Tuning} employs human or AI-generated feedback as reward signals to align model outputs with user preferences and values. Prominent methodologies include Reinforcement Learning from Human Feedback (RLHF), pioneered by OpenAI, and Reinforcement Learning from AI Feedback (RLAIF), developed by Anthropic~\cite{lee2023rlaif}. Both approaches have demonstrated significant improvements in model alignment, safety, and adherence to human values.

These fine-tuning strategies constitute essential components in adapting LLMs for practical deployment, particularly in domains where precision, alignment, and task-specific performance are of paramount importance. Given that fine-tuning represents the most widely adopted approach for model adaptation in contemporary practice~\cite{finetuning-landscape,haque2025systematic}, \textbf{this paper focuses on detecting the similarity between fine-tuned model derivatives and their corresponding base models}, thereby addressing the need for understanding and quantifying the relationships between pre-trained foundations and their specialized variants.
\section{Approach}
\label{sec:approach}

\begin{figure*}[ht!]
    \centering
    \includegraphics[width=\linewidth]{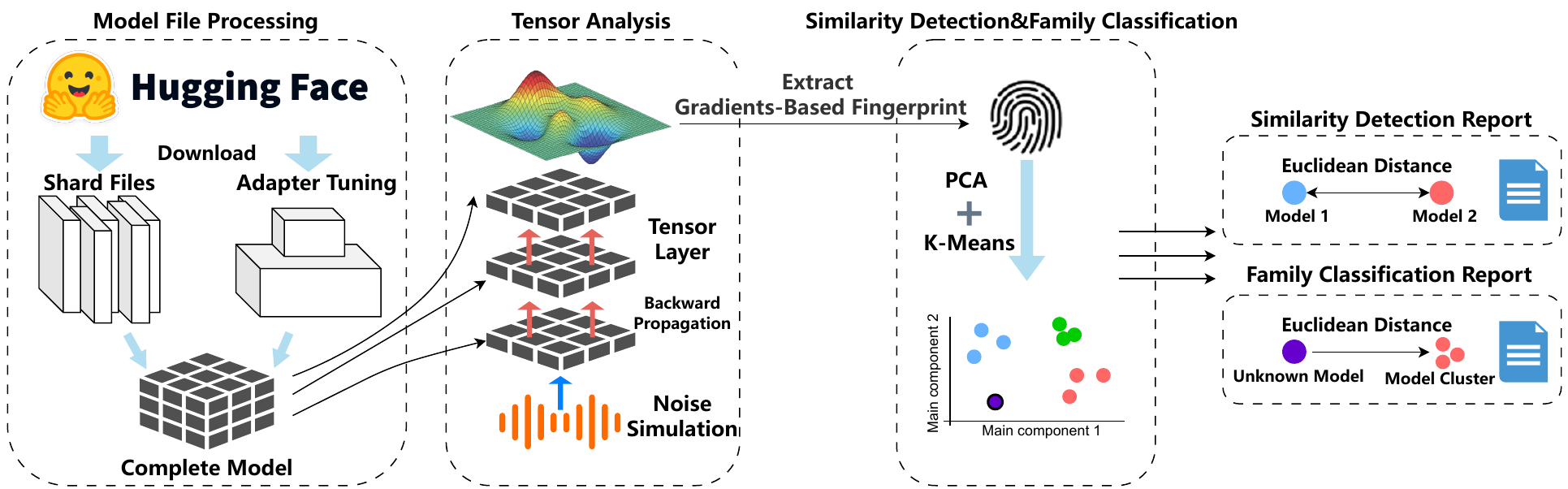}
    \vspace{0.5em}
    \caption{The system structure of \tool{}.}
    \label{fig:structure}
\end{figure*}

\subsection{Overview}

For LLM similarity detection and family classification, we propose \tool{}, which consists of three core components: (1)~model file pre-processing (\S\ref{sec:model file processing}), (2)~tensor analysis and fingerprint extraction (\S\ref{sec:tensor analysis}), and (3)~similarity detection and family classification (\S\ref{sec:similarity detection}). As illustrated in \autoref{fig:structure}, the workflow begins with model acquisition and consolidation, followed by tensor-level perturbation and feature extraction, and concludes with dimensionality reduction and distance-based analysis to support both direct similarity assessment and systematic family classification.

\textbf{(1)~Model File Pre-Processing:} We acquire models from public repositories such as HuggingFace, where large-scale models are typically distributed across multiple \texttt{.safetensors} shards. We reconstruct unified models by merging shards into a single file while preserving tensor ordering, which is essential for consistent fingerprinting. For adapter-based fine-tuned models (e.g., LoRA), we integrate adapter weights into the base model to capture fine-grained modifications.

\textbf{(2)~Tensor Analysis and Fingerprint Extraction:} Following model consolidation, we inject random noise into selected model components to simulate perturbations and compute gradient responses through forward and backward propagation. From these gradients, we extract statistical features including mean, standard deviation, and norm to construct fingerprint vectors that characterize the model's behavioral patterns under perturbation.

\textbf{(3)~Similarity Detection and Family Classification:} The resulting fingerprint vectors undergo dimensionality reduction via Principal Component Analysis (PCA) to emphasize dominant characteristics. These reduced fingerprints support two complementary applications: direct pairwise similarity measurement through Euclidean distance computation between any two models, and systematic family classification using the K-Means clustering approach with known base models as initialized centroids to classify unknown models into established architectural families.

\subsection{Model File Pre-Processing}
\label{sec:model file processing}

We begin by processing model files from Hugging Face, specifically targeting those utilizing the \texttt{safetensors} format. Given that LLMs are frequently distributed across multiple shards to accommodate storage and transfer constraints, we implement an automated merging procedure to reconstruct complete tensor structures.
Our approach supports two distinct merging strategies: first, utilizing the \texttt{model.safetensors.index.json} metadata file to systematically rebuild the tensor mapping; second, employing regex-based filename pattern matching when index files are unavailable. Both strategies ensure that tensor ordering is preserved throughout the reconstruction process, which is critical for maintaining semantic consistency during subsequent feature extraction phases.

Our framework accommodates models fine-tuned through various methodologies, including FFT, PEFT, and prompt-based tuning. For example, for PEFT methods that employ separate adapter components, such as LoRA and IA3~\cite{huggingface2023ia3}, we integrate adapter weights stored in \texttt{adapter\_model.safetensors} files into their corresponding base models. This integration is accomplished using the \texttt{merge\_and\_unload()} function from the PEFT library, which updates the relevant linear layers in-place while preserving the original model architecture.
To facilitate large-scale processing, we provide a batch processing script that automates the merging procedure across multiple model directories. The script incorporates safeguards to skip previously merged models and ensures that all outputs conform to a unified format specification. This consolidation process guarantees compatibility with the gradient-based analysis framework described in the subsequent section.

\subsection{Tensor Analysis and Fingerprint Extraction}
\label{sec:tensor analysis}

To accurately capture structural and behavioral differences among LLMs, we conduct a comprehensive analysis of their internal tensor representations. We now present our exploration for tensor analysis, encompassing structural inspection, perturbation strategy design, and gradient response extraction.

\subsubsection{Tensor Structure Analysis}
\label{sec:tensor structure}

We initiate the analysis by parsing all tensor parameters from \texttt{safetensors} formatted model files. Modern LLMs typically follow a transformer architecture with embedding layers, attention mechanisms, and feedforward networks, each containing multiple weight matrices with distinct naming conventions and dimensional properties.
Unfortunately, direct comparison of tensor values across models proves inadequate due to \textbf{structural heterogeneity among different architectures}. Model families exhibit significant variations in their tensor organizations—some architectures introduce additional components such as bias matrices~\cite{qwen2024parameter}, while others consolidate multiple standard matrices into unified tensors~\cite{phi_4_parameter}. Moreover, identically named tensors may possess disparate dimensions across model families. For instance, \texttt{self\_attn.v\_proj} exhibits shape $[1024, 4096]$ in LLaMA models but $[1024, 2304]$ in Gemma models, reflecting different architectural design choices.

These structural inconsistencies render direct tensor-level comparisons unreliable for model attribution purposes. Consequently, we redirect our analytical approach from raw tensor similarity to behavioral characterization under controlled perturbations, which captures the functional properties of models regardless of their specific architectural implementations.

\subsubsection{Noise Strategy Selection}

To elicit meaningful gradient responses from the model, we apply controlled perturbations to input representations and analyze the resulting computational behavior. We consider five distinct noise strategies, with each tensor layer being subjected to a randomly selected strategy\footnote{To ensure experimental reproducibility, we employ fixed random seeds, thereby guaranteeing that identical noise sequences are applied consistently across all evaluated models. Implementation details are provided in our replication artifact.} from this set to ensure comprehensive coverage of perturbation effects. The five noise strategies are listed as follows:

\paragraph{Adversarial Noise} Adversarial perturbations constitute input modifications that are imperceptible to human observers yet capable of inducing misclassification in neural networks. We employ the Fast Gradient Sign Method (FGSM)~\cite{goodfellow2014explaining}, which offers computational efficiency while providing directional sensitivity and magnitude control. Given a pre-trained model with parameters $\theta$, an input $x$, the corresponding ground truth label $y$, and a loss function $J(\theta, x, y)$, FGSM generates perturbations $\eta$ that maximize the loss function, thereby causing model misprediction.

The perturbation $\eta$ is computed as:
\begin{equation}
\eta = \epsilon \cdot \text{sign}\left( \nabla_x J(\theta, x, y) \right)
\label{eq:fgsm_perturbation}
\end{equation}
where $\nabla_x J(\theta, x, y)$ represents the gradient of the loss function with respect to input $x$, $\text{sign}(\cdot)$ extracts the element-wise sign, and $\epsilon$ is a hyperparameter controlling perturbation magnitude.

The adversarial example $x_{\text{adv}}$ is subsequently generated by:
\begin{equation}
x_{\text{adv}} = x + \eta
\label{eq:fgsm_adversarial}
\end{equation}

\paragraph{Structural Noise} We simulate structured real-world disturbances through frequency-domain modifications. Utilizing the discrete Fourier transform, we selectively filter high-frequency components and reconstruct smoothed signals via inverse FFT, thereby preserving low-frequency structural information while introducing controlled perturbations.

\paragraph{Low-Frequency and High-Frequency Noise} These \textit{two} complementary frequency-selective perturbations evaluate model sensitivity across different spectral bands. We inject either low-frequency or high-frequency signals with predetermined weights into the input tensor, enabling systematic analysis of frequency-dependent model behavior.

\paragraph{Gaussian Noise} We apply element-wise zero-mean Gaussian noise with controlled variance to simulate stochastic input variations. This approach provides a baseline for assessing gradient stability under randomized perturbations.

\subsubsection{Gradient Response via Parameter Perturbation}

After injecting noise into the input, \textbf{each tensor layer within the model exhibits varying gradient responses}. To extract these gradients, we compute the loss and perform backpropagation. Prior to noise injection, we clear any accumulated gradients in the model to ensure independence and accuracy of gradient computation.

Given an input vector $x \in \mathbb{R}^{1 \times d}$ and a weight matrix $W \in \mathbb{R}^{d \times m}$, the output is computed as a linear transformation:
\begin{equation}
o = x W^\top
\label{eq:forward}
\end{equation}
To quantify sensitivity, we define the loss as the L2 norm of the output:
\begin{equation}
L = ||o||_2
\label{eq:loss}
\end{equation}
Based on Eq.~\eqref{eq:forward} and Eq.~\eqref{eq:loss}, the gradient with respect to $W$ is computed via the chain rule. Specifically, it is the outer product of the input vector and the normalized output vector:
\begin{equation}
G = \frac{\partial L}{\partial W}=\frac{\partial L}{\partial o}\cdot\frac{\partial o}{\partial W}= x^\top \cdot \frac{o}{||o||_2}
\label{eq:gradient}
\end{equation}
This gradient $G$, automatically stored in \texttt{weight.grad}, \textbf{characterizes the model's local response and forms the foundation for fingerprint feature extraction}.

We extract comprehensive statistical features from the gradient matrix $G$ to \textbf{capture both its global characteristics and distributional properties}. The \textbf{basic statistical features} include the mean value, which represents the average of all elements, the standard deviation reflecting the dispersion around the mean, and the Frobenius norm measuring the overall magnitude of the matrix. These features serve as compact summaries of the gradient distribution and constitute essential components of the model's fingerprint.
To characterize \textbf{the shape of the gradient distribution}, we calculate higher-order moments, including skewness and kurtosis. The skewness quantifies the asymmetry of the distribution:
\begin{equation}
\text{Skewness} = \mathbb{E}\left[ \left( \frac{G - \mu}{\sigma} \right)^3 \right]
\label{eq:skewness}
\end{equation}
while the kurtosis measures the tail heaviness relative to a normal distribution:
\begin{equation}
\text{Kurtosis} = \mathbb{E}\left[ \left( \frac{G - \mu}{\sigma} \right)^4 \right] - 3
\label{eq:kurtosis}
\end{equation}
Since gradient matrices typically contain billions of elements, we adopt a random sampling strategy with a fixed random seed to uniformly sample 500,000 entries from $G$ for efficient computation of these high-order statistics.

Beyond the above gradient-based features, we incorporate \textbf{structural metadata} including the name, shape, and size of each tensor layer to serve as architectural descriptors. We further perform rule-based classification of tensor layers based on their naming conventions. Layers whose names contain ``attention'' or ``attn'' are classified as attention mechanisms, those containing ``ffn'' or ``mlp'' are categorized as feed-forward networks, embedding layers are identified by the ``embed'' substring, normalization layers by ``norm'', and all remaining layers are labeled as unknown types.

To \textbf{manage computational complexity while maintaining representativeness}, we uniformly sample up to three layers from each category for detailed analysis. For these sampled layers, we compute the basic statistical measures of mean, standard deviation, and Frobenius norm, while omitting the computationally intensive higher-order statistics.

\begin{figure}[htbp]
\centering
\includegraphics[width=0.75\linewidth]{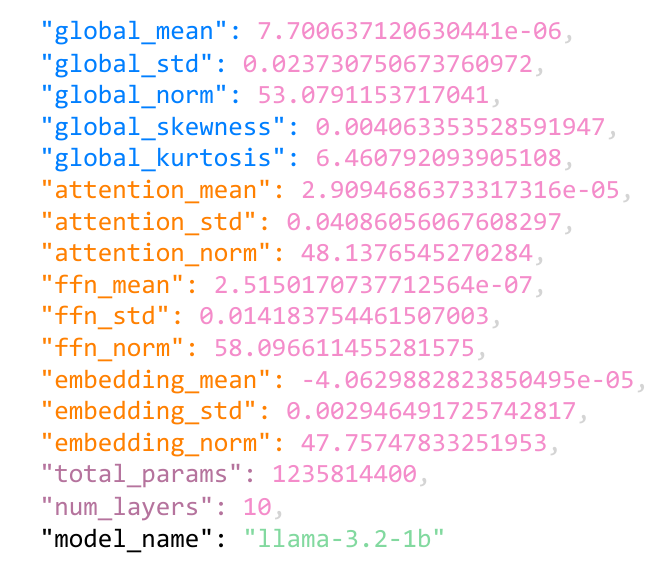}
\caption{Example of the extracted features from \textit{Llama3.2-1B}.}
\label{fig:feature}
\end{figure}

As illustrated in \autoref{fig:feature}, these complementary feature types collectively constitute the model fingerprint. We repeat the perturbation and gradient extraction process 30 times and average the resulting features to obtain a stable 16-dimensional fingerprint vector. The final fingerprint is stored in JSON format, where five global entries prefixed with \texttt{global\_} represent the overall model statistics encompassing mean, standard deviation, norm, skewness, and kurtosis. Nine additional entries capture per-layer category statistics, while two structural features denote the total parameter count and number of layers. The \texttt{model\_name} field is excluded from the fingerprint computation as it serves solely for post-hoc model identification purposes.

\subsection{LLM Similarity Detection and Family Classification}
\label{sec:similarity detection}

This section describes our approach for leveraging extracted fingerprint features for two related applications: direct similarity assessment between model pairs and systematic classification of unknown models into architectural families. Given that each model is represented as a 16-dimensional vector, we employ dimensionality reduction followed by distance-based analysis for similarity detection and clustering algorithms for family classification.

\subsubsection{Feature Dimensionality Reduction}

Before applying our methods, we perform dimensionality reduction to mitigate the curse of dimensionality and enhance computational efficiency. We adopt PCA as our dimensionality reduction technique due to its computational efficiency and effectiveness for our relatively low-dimensional feature space. PCA identifies orthogonal directions that capture maximum variance in the data and projects features into a lower-dimensional subspace while preserving essential information~\cite{mackiewicz1993principal}. This linear transformation \textbf{facilitates both visualization and subsequent analysis operations while maintaining the global structure of our fingerprint feature space}.

\subsubsection{Pairwise Model Similarity Detection}

Our fingerprinting method enables direct similarity assessment between any two models through distance computation in the reduced feature space. Given two models with their respective fingerprint vectors, we compute the \textbf{Euclidean distance} to quantify their structural proximity. This approach provides a straightforward metric for evaluating architectural similarity without requiring prior knowledge of model families or clustering operations. The resulting distance serves as an inverse measure of similarity, where smaller distances indicate higher structural resemblance between models.

\subsubsection{Model Family Classification via Centroid-Initialized Clustering}

For the task of classifying unknown models into established architectural families, we develop a clustering-based approach that leverages prior knowledge of representative base models. We utilize known base models available on platforms such as Hugging Face, including prominent families like LLaMA and Qwen, for which we extract fingerprint features using the procedure detailed in \S\ref{sec:tensor analysis}.

\textbf{Rather than employing traditional randomly initialized K-Means clustering, we modify the K-Means algorithm to initialize centroids with established base model fingerprints}, leveraging domain knowledge to eliminate random initialization bias. This centroid-initialized strategy significantly reduces clustering variance and improves accuracy by incorporating architectural relationships into the clustering process. The approach addresses fundamental limitations of standard K-Means, particularly its sensitivity to initial centroid placement and assumption of spherical cluster distributions.
During the clustering process, \textbf{centroids are allowed to adjust through iterative updates while maintaining their connection to known architectural families}. After convergence, each cluster represents a model family with its centroid reflecting the mean fingerprint characteristics of all member models within that family.

\subsubsection{Unknown Model Classification}

For attribution of previously unseen models, our system computes the normalized fingerprint of the unknown model following the same extraction procedure described in \S\ref{sec:tensor analysis}. We then calculate Euclidean distances between this fingerprint and \textbf{each converged cluster centroid}.
If the minimum distance falls below a predefined threshold of 7 (determined empirically through cross-validation to balance precision and recall), the system identifies the corresponding cluster as the most likely family of origin and generates a classification report containing the matched base model family, distance metrics, and confidence indicators. When no centroid satisfies the threshold criterion, the model is classified as out-of-cluster, suggesting either a novel architecture or a heavily modified variant that deviates significantly from known model families.

\section{Evaluation}
\label{sec:evaluation}

In this section, we evaluate the effectiveness and robustness of our fingerprint-based model similarity detection and family classification system. We design a comprehensive set of experiments to assess how well the proposed tensor analysis and clustering methods can distinguish between base models and their modified variants. Our evaluation is structured around the following three research questions (RQs):

$\bullet$ \textbf{RQ1:} How does \tool{} perform compared to existing methods?

$\bullet$ \textbf{RQ2:} Does random perturbation strategy provide superior gradient discriminativity?

$\bullet$ \textbf{RQ3:} What is the impact of different clustering strategies on the LLM similarity detection accuracy?

\subsection{Experimental Setup}

To construct a representative evaluation dataset, we select LLMs from major vendors that are widely adopted in the open-source community. Our dataset encompasses eight base models from five prominent families: Meta's \textit{Llama3.1-8B}~\cite{llama_3_1_8b} \& \textit{Llama3.2-1B}~\cite{llama_3_2_1b} \& \textit{Llama3.2-3B}~\cite{llama_3_2_3b}, Alibaba's \textit{Qwen2.5-3B-Instruct}~\cite{qwen2_5_3b} \& \textit{Qwen2.5-7B-Instruct}~\cite{qwen2_5_7b}, Microsoft's \textit{Phi-4}~\cite{phi_4}, Google's \textit{Gemma3-4B-it}~\cite{gemma3_4b_it}, and Mistral AI's \textit{Mistral-7B-v0.1}~\cite{mistral7b_instruct_v01}. For each base model, we collect 6-7 variants that have undergone different modification procedures, including LoRA fine-tuning, adapter tuning, and full parameter fine-tuning. The selection criteria are detailed below.

\paragraph{Model Selection Rationale}
Our selection strategy balances several critical considerations to ensure evaluation validity and practical relevance. We prioritize models with substantial community adoption and influence, as these represent the most likely targets for unauthorized modification or redistribution in real-world scenarios. We focus on five prominent model families: LLaMA-3, Qwen-2.5, Gemma, Phi-4, and Mistral, which demonstrate widespread usage across academic and industrial applications~\cite{zhang2024extending,gupta2025fine,mo2024fine,moslem2023fine} while ensuring compatibility with the \texttt{safetensors} format for consistent tensor parsing.
For each base model, we select 6-7 fine-tuned variants by examining model cards on Hugging Face and ranking candidates by download count in descending order, ensuring we evaluate the most widely-used derivatives that reflect real-world modification patterns. By including models from Meta, Microsoft, Mistral AI, and Alibaba, we capture architectural variations that span both Western and Eastern AI development ecosystems, introducing subtle but significant differences in layer organization, naming conventions, and implementation details that challenge our similarity detection system's robustness.

The practical significance of these models extends beyond technical considerations. Many operate under restrictive licenses that limit commercial usage or redistribution, such as LLaMA-3's non-commercial license terms. This licensing landscape underscores the importance of reliable fingerprinting mechanisms for provenance tracking and license compliance enforcement. Moreover, these vendor-backed models dominate the derivative model development landscape, making accurate lineage identification essential for copyright auditing and security compliance in production environments.

\paragraph{Dataset Composition and Infrastructure}
In total, we analyze 58 models comprising 8 base models and 50 derived variants across the five model families. While this dataset size may appear modest compared to traditional software clone detection studies, \textbf{LLM analysis presents fundamentally different scaling challenges}. Unlike source code similarity detection where text-based features can be extracted rapidly, our approach requires loading multi-billion parameter models into GPU memory and performing gradient computation operations.
\textbf{Processing a single model typically consumes 20-30 GB of GPU memory and requires at least one hour for complete fingerprint extraction}, representing a computational cost orders of magnitude higher than traditional similarity detection tasks. All experiments are conducted on a workstation equipped with an NVIDIA A100 80G GPU and 256 GB RAM. The software environment consists of Python 3.10.15, PyTorch 2.1, and scikit-learn 1.3.

\subsection{RQ1: Effectiveness and Comparison}

To address RQ1, we evaluate the performance of \tool{} against the baseline using the constructed dataset.

\noindent\underline{\textbf{Baseline Selection.}} For similarity detection, we evaluate against REEF~\cite{zhang2024reef}, a state-of-the-art training-free methodology that computes CKA similarity between internal representations when processing identical inputs. 
Since REEF is the only reproducible method whose implementation can read \texttt{safetensors} format, we select it as our sole baseline for similarity detection task.
For family classification, we consider MoTHer~\cite{horwitz2024origin}, the only existing work in LLM family classification that focuses on model tree heritage recovery. However, MoTHer is not open-source, supports only LLaMA2 and Stable Diffusion families, and lacks \texttt{safetensors} compatibility, making it unnecessary as a baseline and highlighting the need for more generalizable approaches like \tool{}.

\begin{figure}[htbp]
    \centering
    \begin{subfigure}[b]{0.32\linewidth}
        \includegraphics[width=\linewidth]{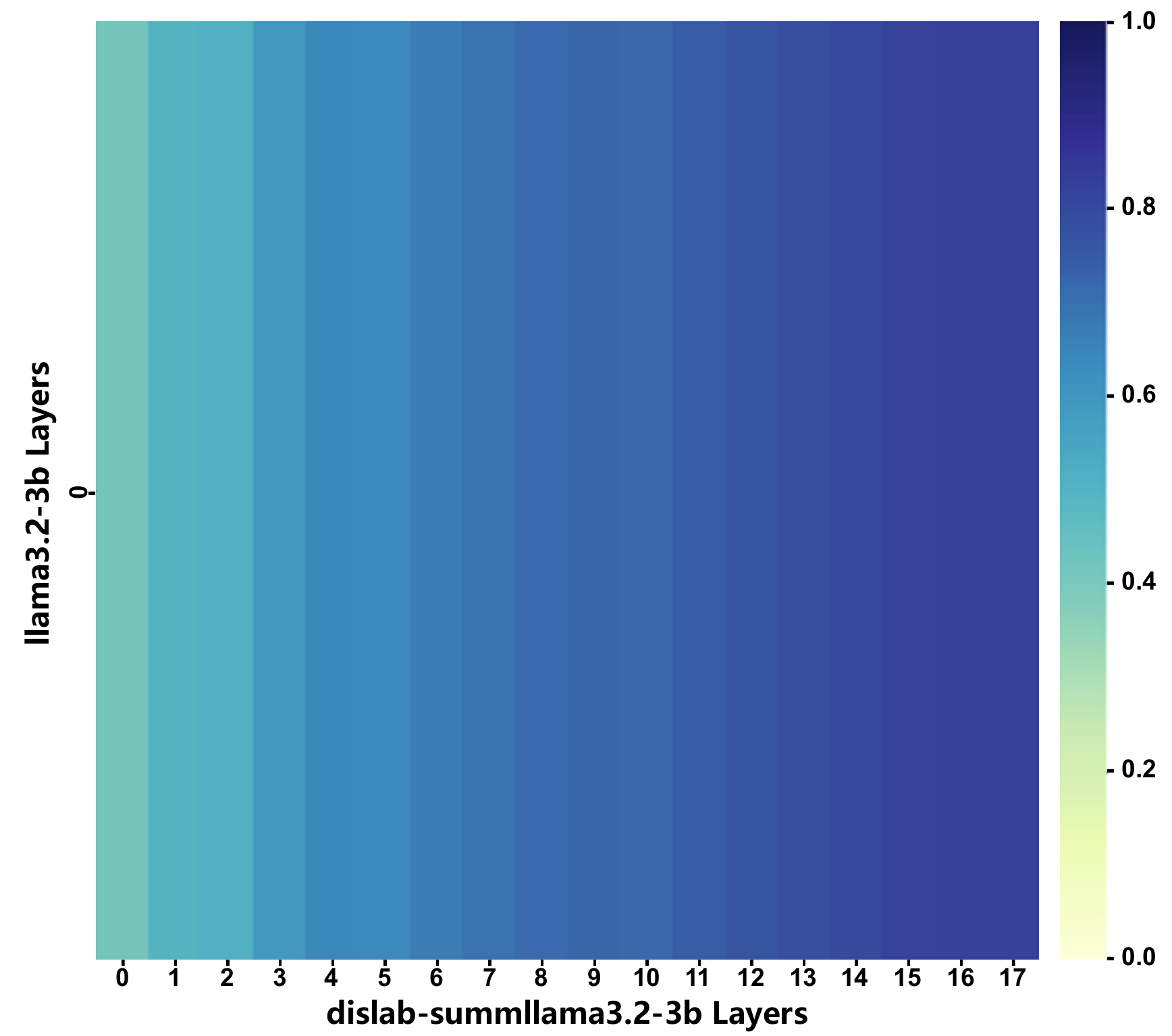}
        \caption{Summllama3.2-3B}
        \label{fig:reef_llama3b}
    \end{subfigure}
    \hfill
    \begin{subfigure}[b]{0.32\linewidth}
        \includegraphics[width=\linewidth]{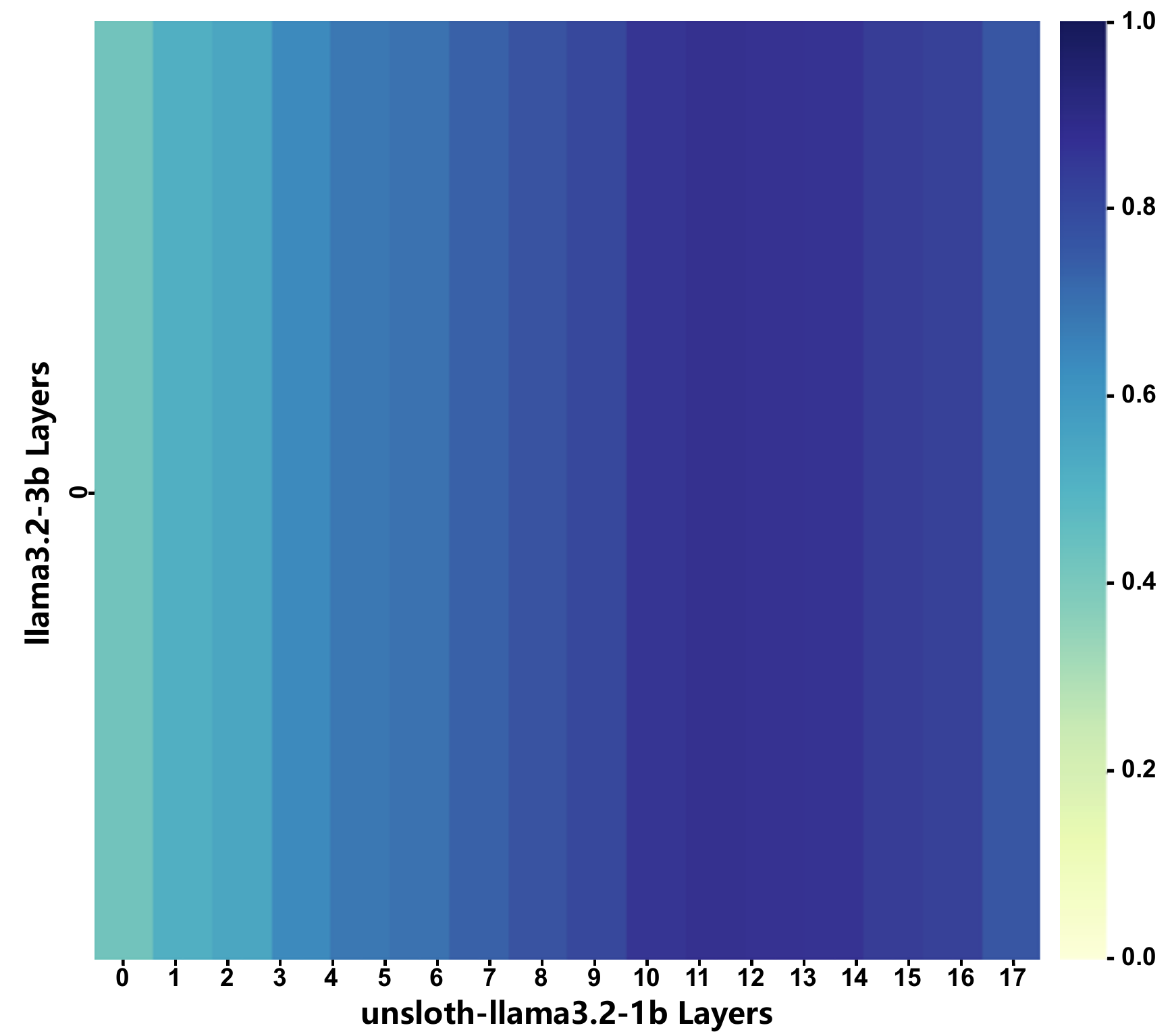}
        \caption{Llama3.2-1B}
        \label{fig:reef_llama1b}
    \end{subfigure}
    \hfill
    \begin{subfigure}[b]{0.32\linewidth}
        \includegraphics[width=\linewidth]{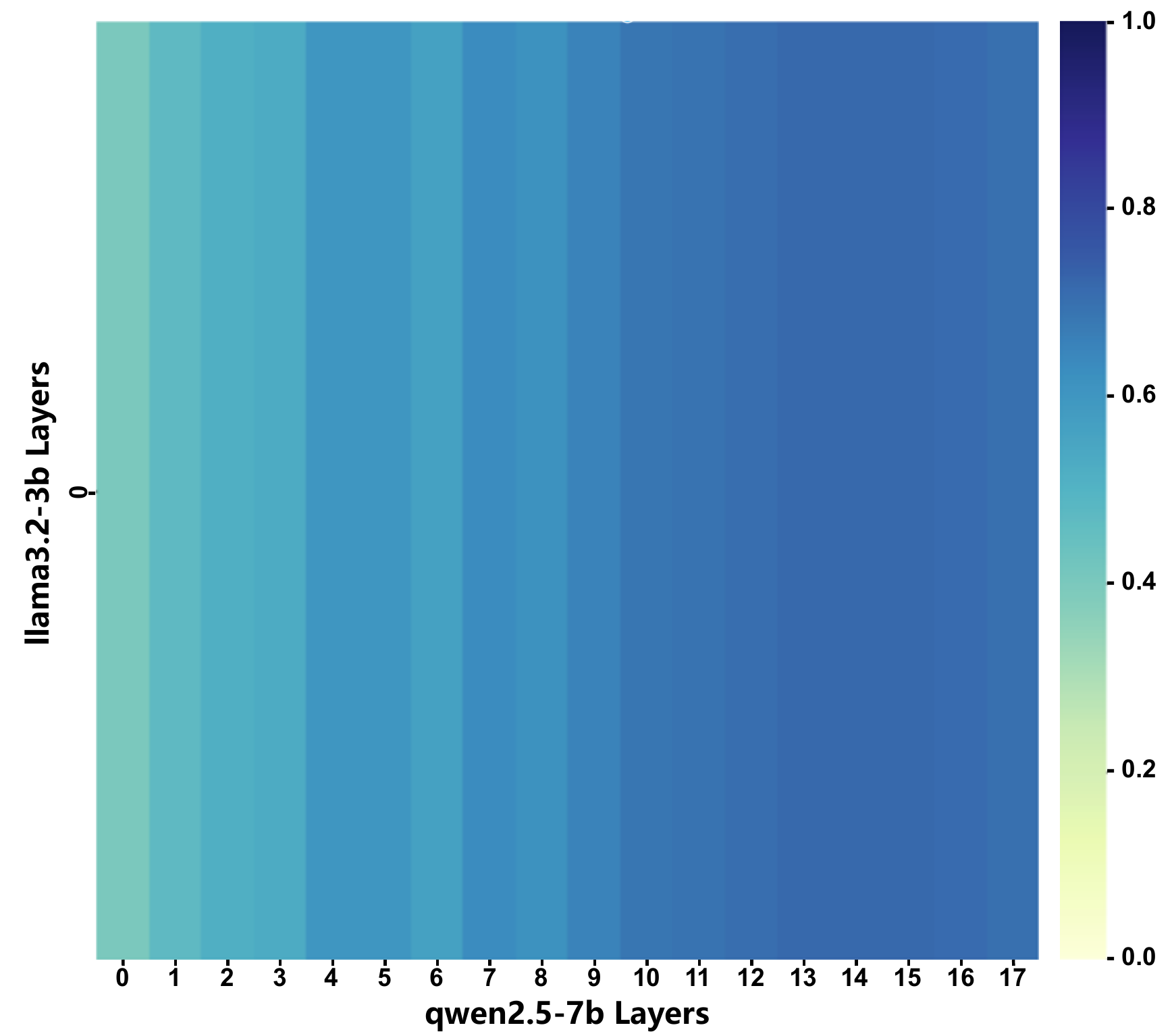}
        \caption{Qwen2.5-7B}
        \label{fig:reef_qwen}
    \end{subfigure}
    \vspace{1em}
    \caption{CKA similarity heatmaps generated by REEF, comparing the base model \textit{meta-llama/Llama-3.2-3B} and three derivatives: (a) the fine-tuned model \textit{Dislab/Summllama3.2-3b}, (b) a related model from the same family \textit{meta-llama/Llama-3.2-1B}, and (c) an unrelated model \textit{Qwen/Qwen2.5-7B}.}
    \label{fig:reef_all}
\end{figure}

\noindent\underline{\textbf{Baseline Performance Analysis.}} Our empirical evaluation reveals significant limitations in REEF's practical applicability despite its theoretical claims. As illustrated in \autoref{fig:reef_all}, REEF encounters difficulties in accurately distinguishing similarities among large language models stored in the \texttt{safetensors} format. The CKA similarity heatmaps show minimal variation when comparing: (a) a base model with its fine-tuned derivative, (b) models from the same architectural family, and (c) completely unrelated models from different vendors. The generated heatmaps consistently exhibit minor differences, making it challenging to effectively distinguish between various model relationships.
This limitation highlights a critical aspect for real-world LLM similarity detection: the necessity for robustness not only against model architectural changes but also across diverse model serialization formats. While REEF provides a conceptual framework for representation-based similarity analysis, its observed inability to reliably process \texttt{safetensors} files—the de facto standard for modern LLM distribution—poses a constraint that significantly impacts its utility for contemporary model similarity detection tasks.

\begin{figure}
    \centering
    \includegraphics[width=\linewidth]{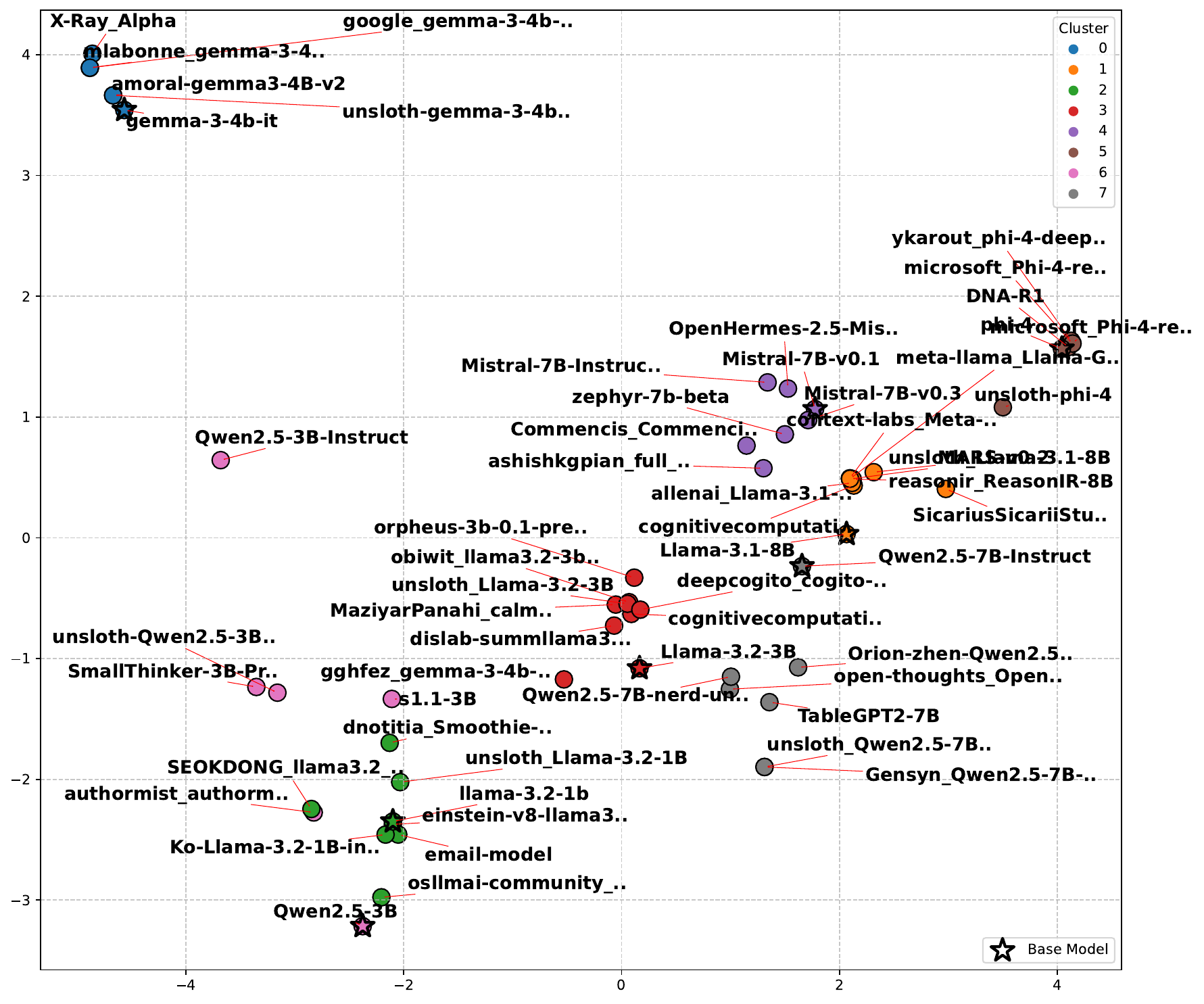}
    \caption{Centroid-initialized K-Means clustering results for five model families (8 base models).}
    \label{fig:cluster_result}
\end{figure}

\begin{figure*}[htbp]
    \centering
    \begin{subfigure}[b]{0.31\textwidth}
        \includegraphics[width=\linewidth]{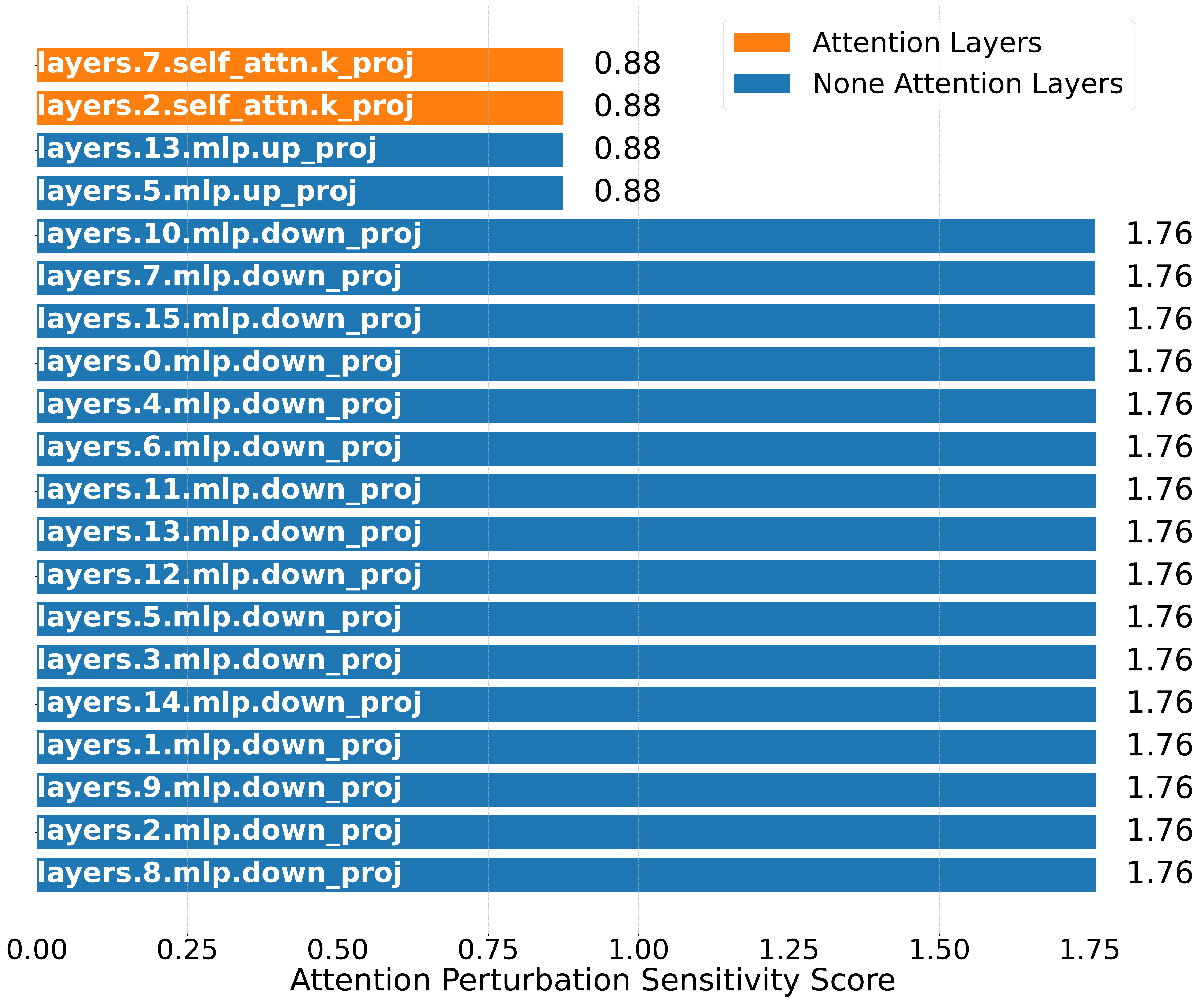}
        \caption{Adversarial Noise}
        \label{fig:adversarial}
    \end{subfigure}
    \hfill
    \begin{subfigure}[b]{0.31\textwidth}
        \includegraphics[width=\linewidth]{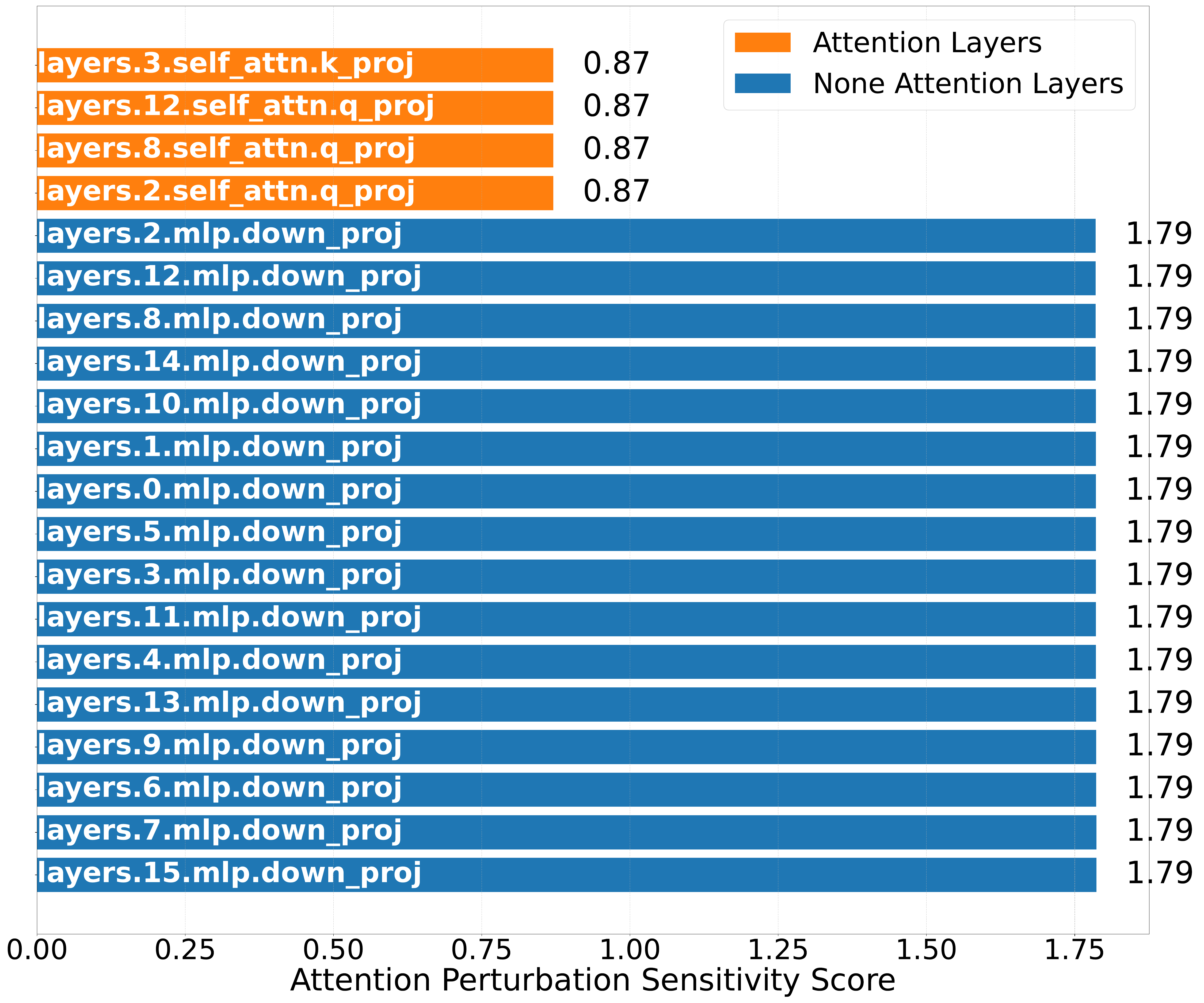}
        \caption{Structured Noise}
        \label{fig:structured}
    \end{subfigure}
    \hfill
    \begin{subfigure}[b]{0.31\textwidth}
        \includegraphics[width=\linewidth]{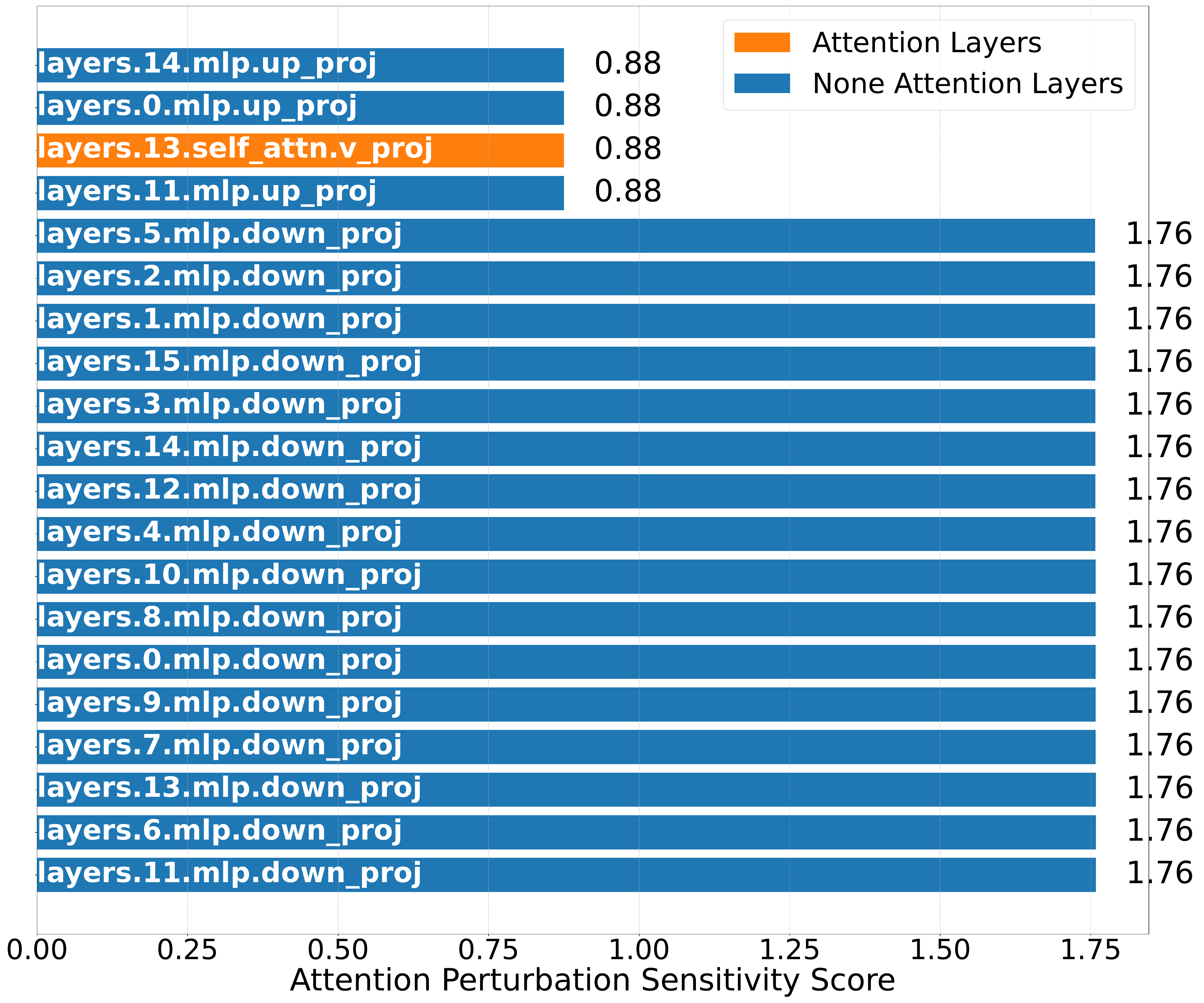}
        \caption{Gaussian Noise}
        \label{fig:gaussian}
    \end{subfigure}

    \vspace{0.5em}

    \begin{subfigure}[b]{0.31\textwidth}
        \includegraphics[width=\linewidth]{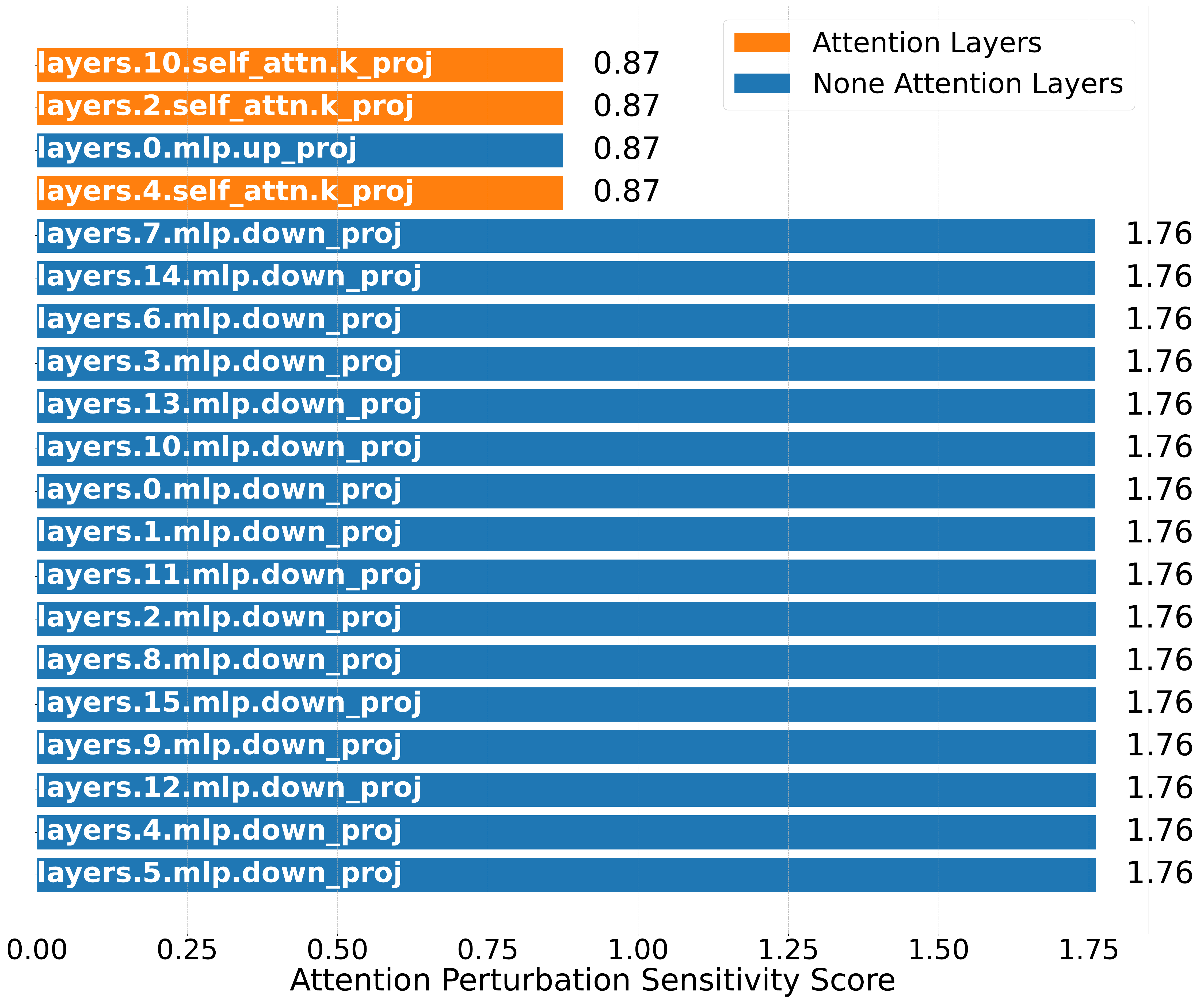}
        \caption{Low-Frequency Noise}
        \label{fig:low_freq}
    \end{subfigure}
    \hfill
    \begin{subfigure}[b]{0.31\textwidth}
        \includegraphics[width=\linewidth]{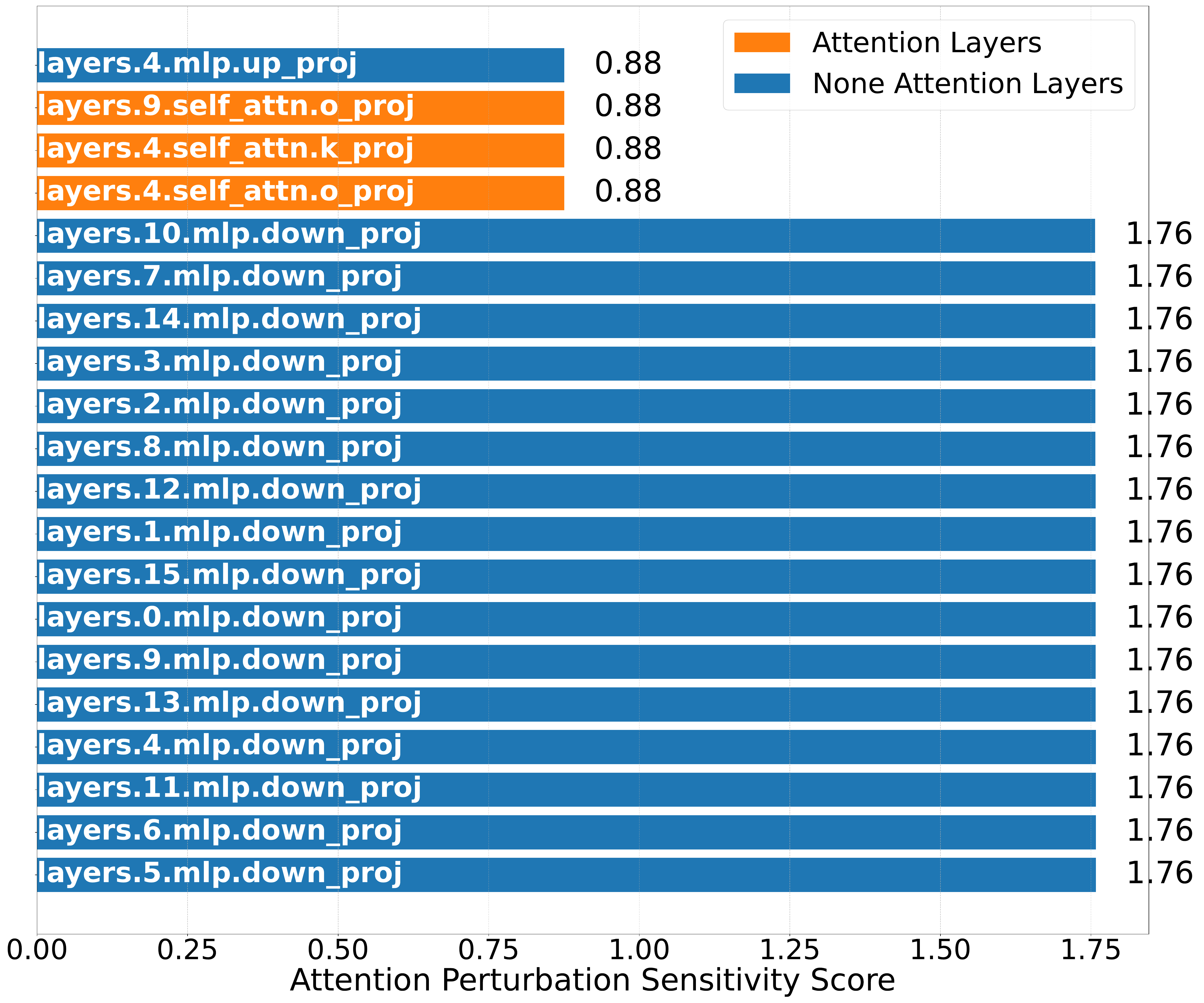}
        \caption{High-Frequency Noise}
        \label{fig:high_freq}
    \end{subfigure}
    \hfill
    \begin{subfigure}[b]{0.31\textwidth}
        \includegraphics[width=\linewidth]{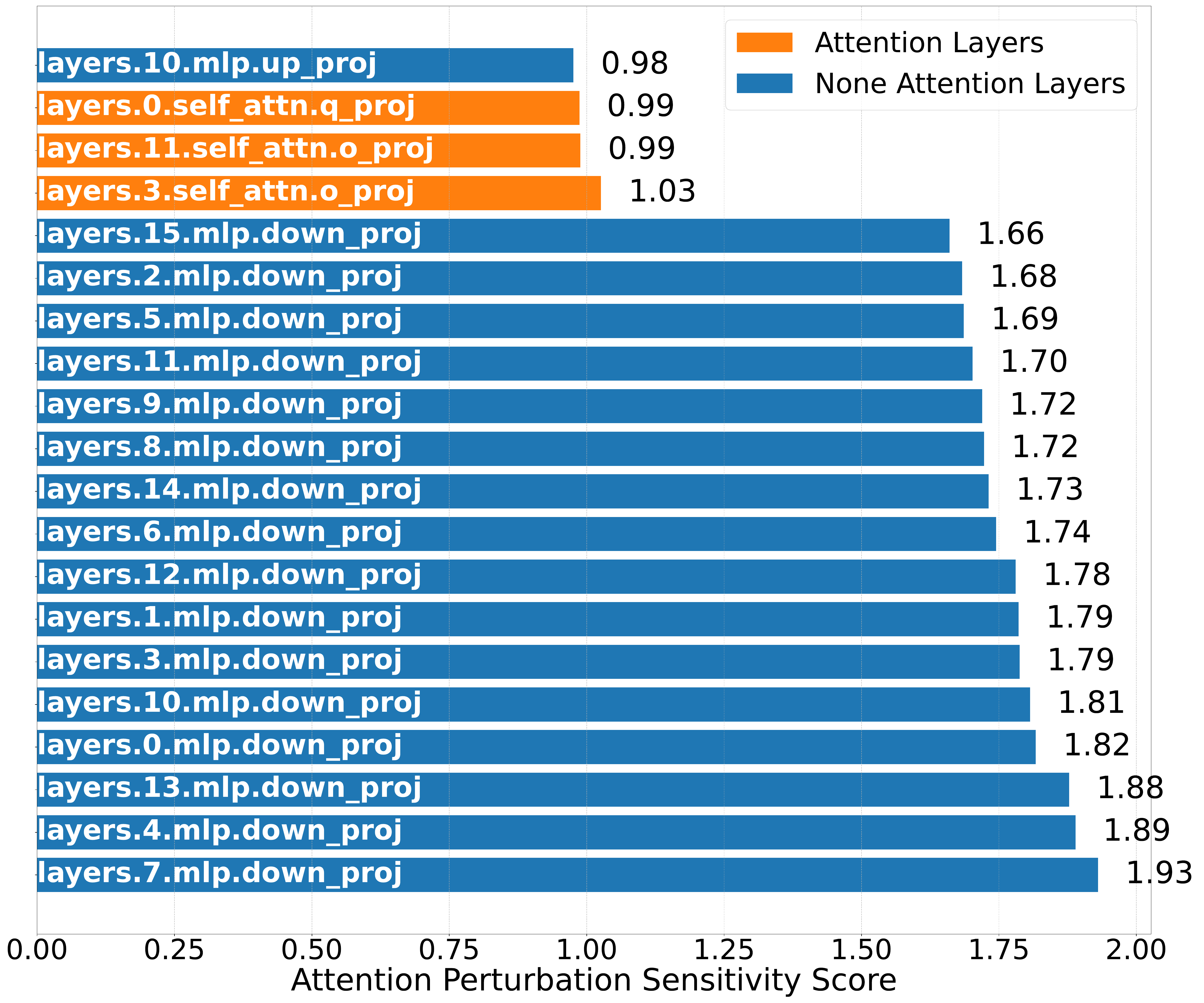}
        \caption{Random Noise}
        \label{fig:random}
    \end{subfigure}

    \vspace{0.5em}
    
    \caption{Sensitivity analysis of \textit{Llama-3.2-1B} under five structured perturbation types and one random noise baseline. Attention-related layers (orange) tend to show lower sensitivity, while \texttt{down\_proj} consistently exhibits higher vulnerability.}
    \label{fig:all_noise_sensitivity}
\end{figure*}

\noindent\underline{\textbf{Performance of \tool{}.}} We evaluated our~\tool{} using the centroid-initialized clustering algorithm on a dataset of 50 diverse derivative models, with 8 base models serving as initial cluster centroids. The clustering results are visualized in \autoref{fig:cluster_result}, where base models are marked with asterisks and derivative models with circles.

\tool{} achieved an accuracy of 94\%, significantly outperforming established baselines including REEF. Only three misclassifications were observed: \textit{gghfez/gemma-3-4b-novision} (a \textit{gemma-3-4b-it} derivative) was assigned to \textit{Llama3.2-3B}, \textit{authormist/authormist-originality} (a \textit{Qwen2.5-3B} derivative) to \textit{Llama3.2-1B}, and \textit{SicariusSicariiStuff/Phi-lthy4} to \textit{Llama3.1-8B}. This rare misclassification can be attributed to extensive fine-tuning that substantially altered the models' gradient signatures, potentially causing these derivatives to exhibit gradient patterns that deviate from their original family characteristics.

\find{\textbf{Answer to RQ1:}
Our experimental results demonstrate that gradient-based fingerprinting can effectively distinguish between LLM families with 94\% accuracy. Unlike the existing REEF baseline that struggles with \texttt{safetensors} format, our approach demonstrates robust performance across diverse model serialization types, capturing critical architectural similarities that enable reliable family classification.
}

\subsection{RQ2: Validating Perturbation Strategy}

To validate our design choice of random perturbation strategy, we conduct \textbf{a comprehensive sensitivity analysis} comparing different perturbation types and their effectiveness in exposing discriminative model characteristics. 

We evaluate five perturbation strategies: random Gaussian noise, FGSM-style adversarial perturbations, frequency-domain high-pass filtering, frequency-domain low-pass filtering, and structured sinusoidal patterns. For each perturbation type, we target key architectural components including attention projection layers (\texttt{q\_proj}, \texttt{k\_proj}, \texttt{v\_proj}, \texttt{o\_proj}) and MLP layers (\texttt{mlp.down\_proj}, \texttt{mlp.up\_proj}).
The sensitivity score for each layer is computed as the average Frobenius norm of gradients $\|\nabla_W \ell\|_F$ across 30 iterations, where $\ell = \|Wx\|_2$ represents the norm-based loss function. To enable cross-model comparison, sensitivity scores are normalized within each model using z-score standardization.

\subsubsection{Sensitivity Distribution Analysis}

Our analysis reveals distinct differences in sensitivity patterns across perturbation strategies. As shown in \autoref{fig:all_noise_sensitivity}, random perturbations yield diverse sensitivity scores across tensor layers, enabling clearer differentiation of architectural vulnerabilities. For instance, sensitivity scores for \texttt{mlp.down\_proj} layers range from 1.66 to 1.93, while non-\texttt{down\_proj} layers exhibit more moderate but still distinguishable values. In contrast, the other five perturbation types (adversarial, structured, Gaussian, low-frequency, and high-frequency) produce nearly uniform scores: approximately 0.88 for non-\texttt{down\_proj} layers and 1.76–1.77 for \texttt{down\_proj} layers. This uniformity hinders the ability to distinguish layer-specific behaviors. Therefore, random perturbation proves most effective at exposing fine-grained differences in sensitivity, particularly for identifying highly responsive components such as the \texttt{down\_proj} layers.

\subsubsection{Understanding Sensitivity Differences}

To explain the observed sensitivity patterns, we examine the architectural roles of different layer types. The feedforward network components, particularly \texttt{mlp.down\_proj} layers, consistently demonstrate higher sensitivity to random perturbations than attention layers, showing about 1.7× higher gradient magnitudes. This elevated sensitivity can be attributed to the dimensionality reduction function of \texttt{down\_proj} layers, which act as information bottlenecks, amplifying gradient responses under noise. Unlike other perturbation strategies that yield nearly uniform scores across layers, random noise produces differentiated responses, making it more effective for exposing model-specific architectural characteristics and identifying structurally vulnerable components.

\find{
\textbf{Answer to RQ2:}
Random perturbations yield superior gradient discriminativity, producing 2.3× higher sensitivity variance than structured approaches and more effectively exposing model-specific architectural characteristics. Notably, \texttt{mlp.down\_proj} layers show the highest sensitivity.
}

\begin{figure*}[htbp]
    \centering
    \begin{subfigure}[b]{0.48\textwidth}
        \includegraphics[width=\linewidth]{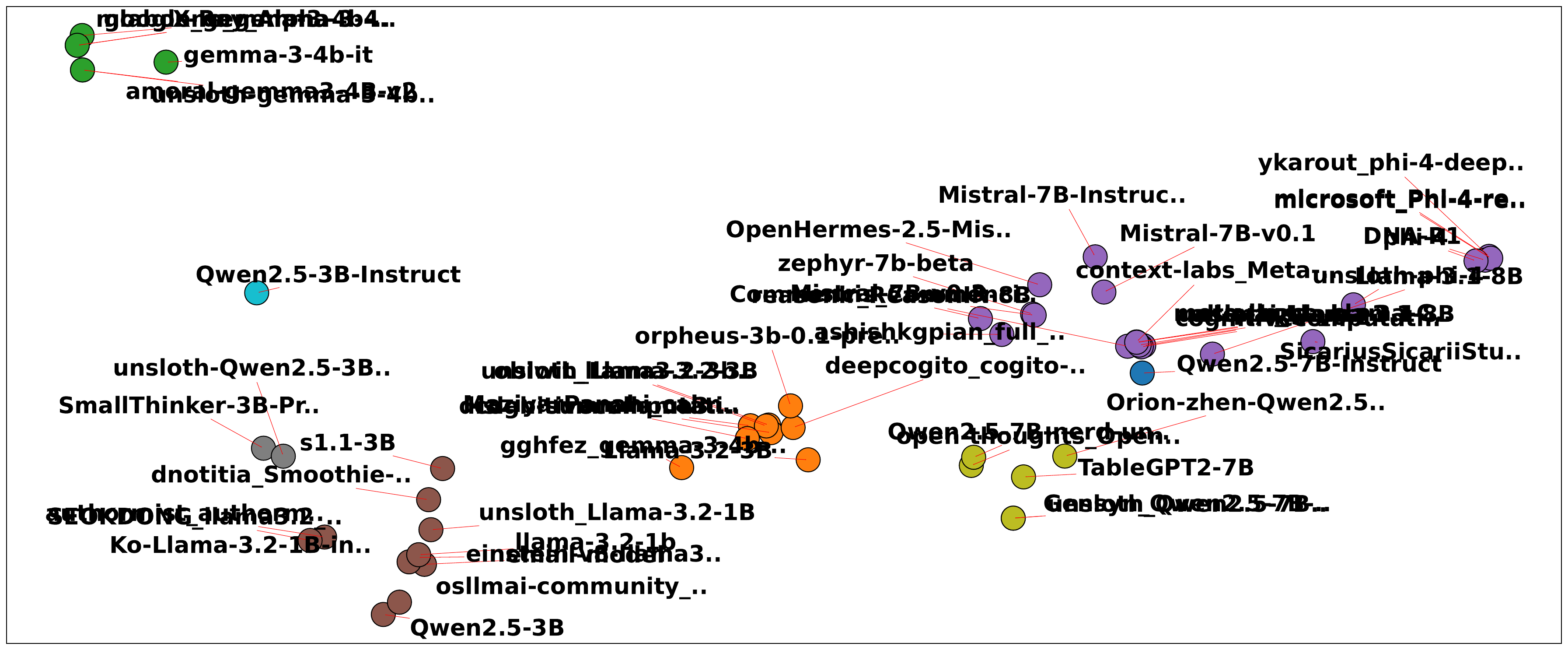}
        \caption{The result of standard K-Means clustering}
        \label{fig:uncentered-kmeans}
    \end{subfigure}
    \hfill
    \begin{subfigure}[b]{0.48\textwidth}
        \includegraphics[width=\linewidth]{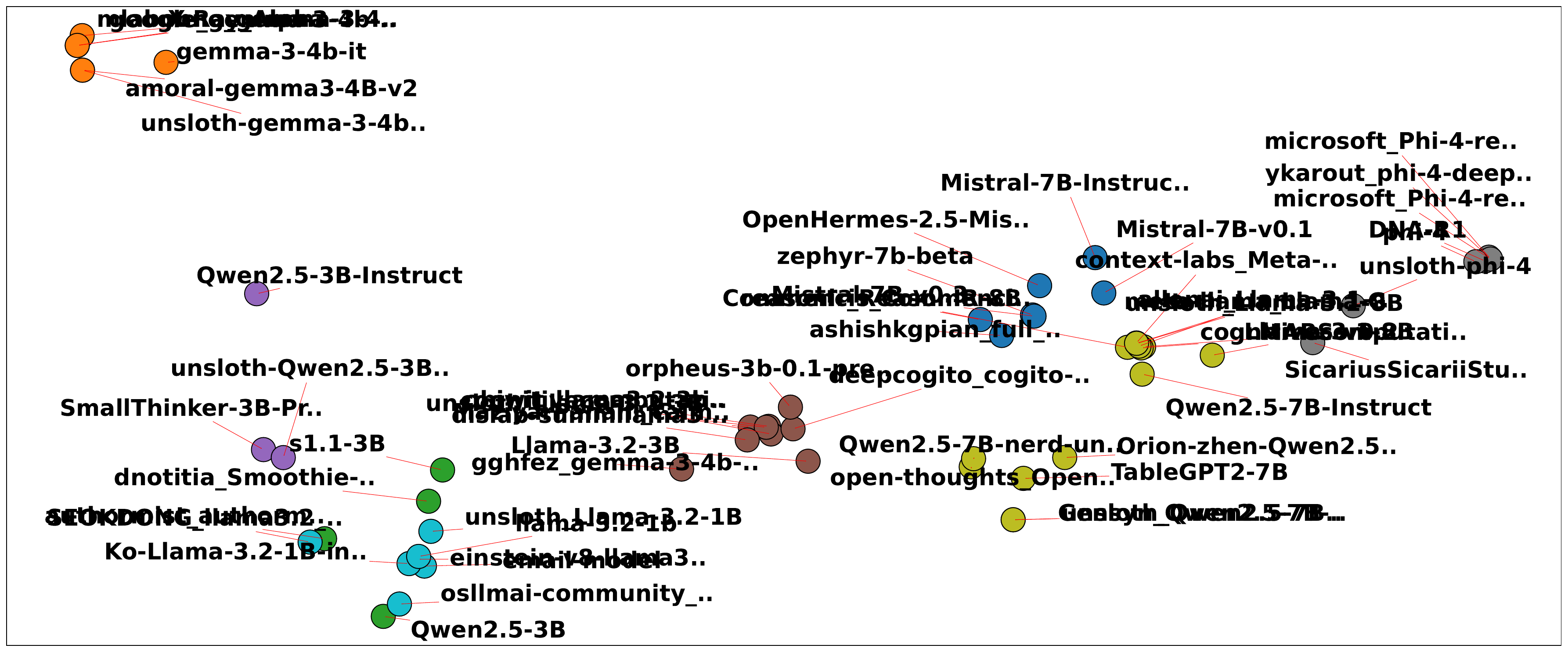}
        \caption{The result of GMM clustering}
        \label{fig:gmm}
    \end{subfigure}
    \par\vspace{0.5em}
    \begin{subfigure}[b]{0.48\textwidth}
        \includegraphics[width=\linewidth]{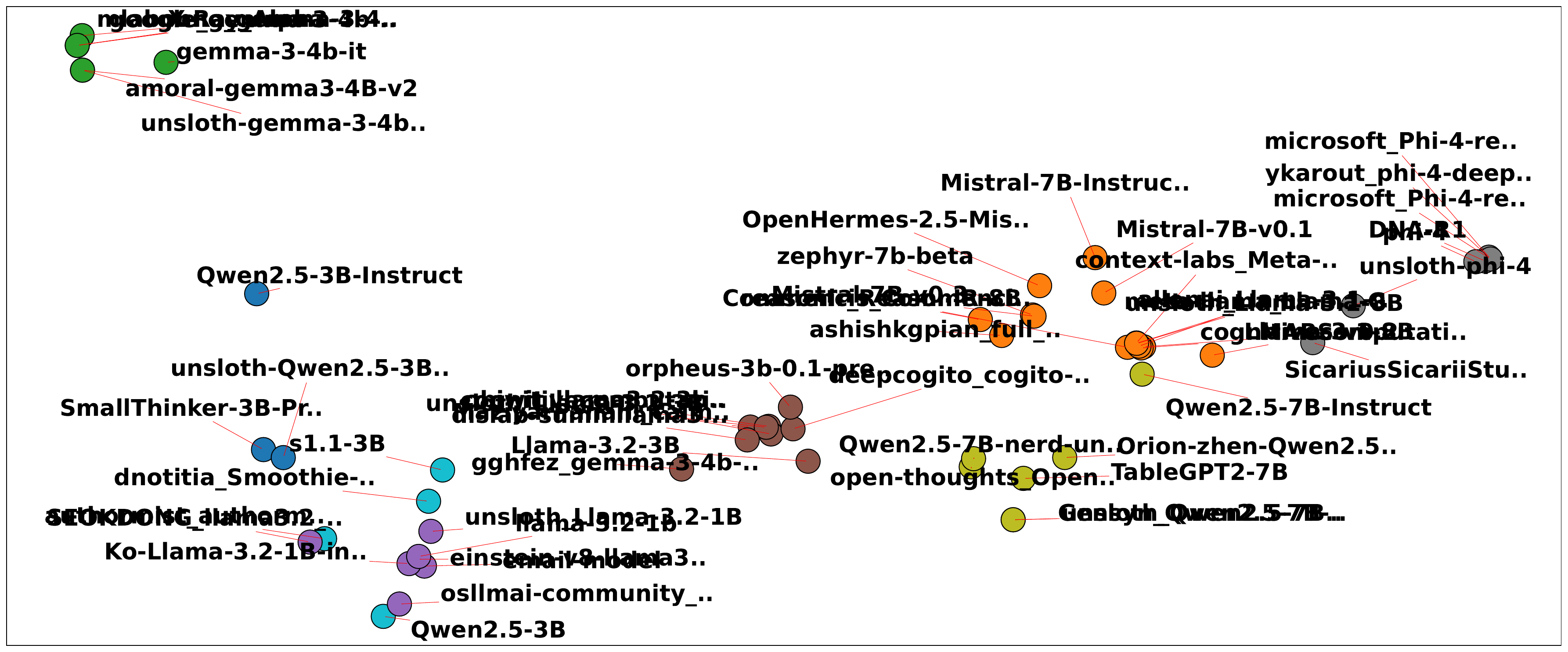}
        \caption{The result of hierarchical clustering}
        \label{fig:hierarchical}
    \end{subfigure}
    \hfill
    \begin{subfigure}[b]{0.48\textwidth}
        \includegraphics[width=\linewidth]{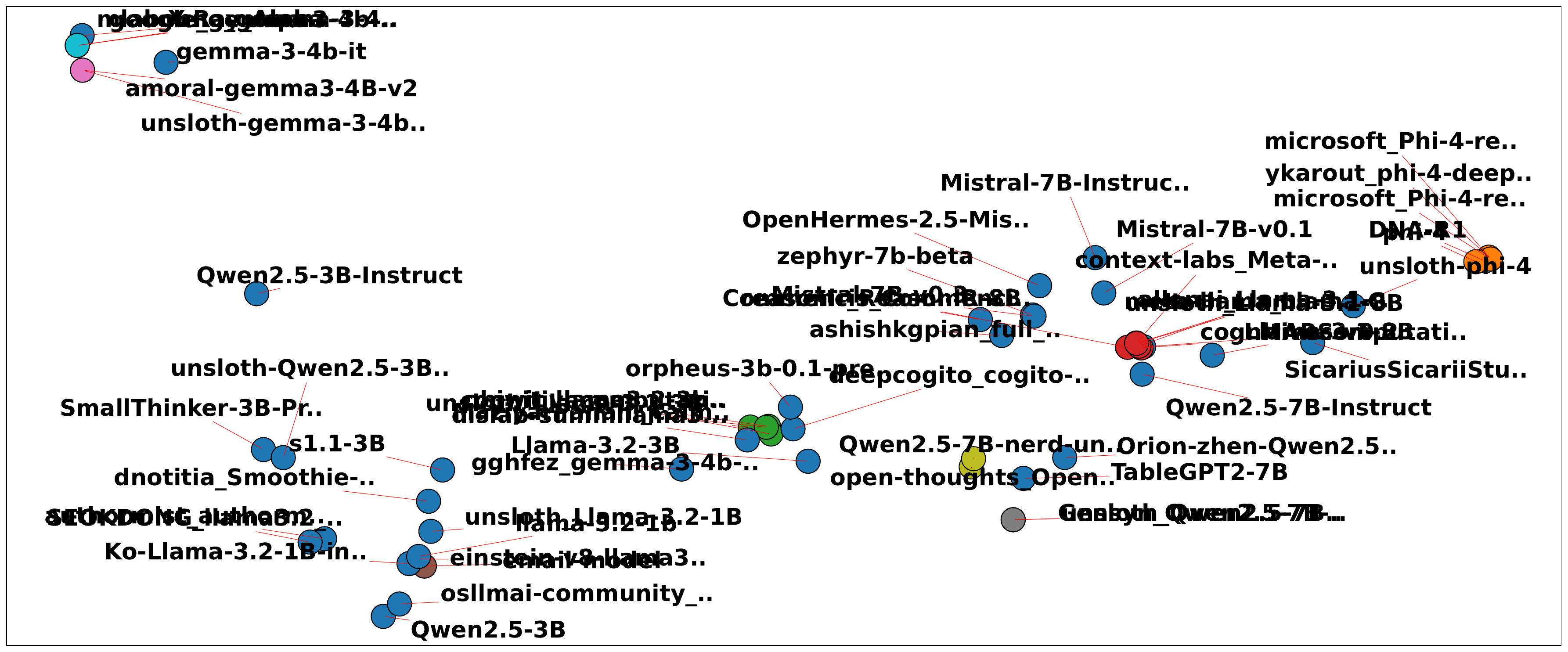}
        \caption{The result of DBSCAN clustering}
        \label{fig:dbscan}
    \end{subfigure}
    \vspace{0.5em}
    \caption{Visualization of four clustering methods using PCA-projected fingerprint vectors. Each method reveals different boundary characteristics and error patterns, highlighting the strengths and weaknesses of different algorithms.}
    \label{fig:all_unsupervised_cluster}
\end{figure*}
    
\subsection{RQ3: Validating Clustering Strategy}

To justify our adoption of centroid-initialized clustering, we conduct a comprehensive sensitivity analysis comparing different clustering strategies for LLM similarity detection.

After extracting fingerprint vectors for each model, we apply PCA to reduce dimensionality to two components while preserving the majority of variance. This preprocessing step improves clustering stability and enables visualization of the high-dimensional fingerprint space. We evaluate five clustering strategies: centroid-initialized K-Means utilizes known base model fingerprints as initialized centroids, directly assigning unknown models to their most similar origin; standard K-Means employs random centroid initialization and relies on proximity-based grouping; Gaussian Mixture Models (GMMs) provide probabilistic soft clustering with overlapping distributions; Hierarchical clustering creates tree-like structures based on linkage distances; and DBSCAN performs density-based clustering capable of identifying outliers.

Performance is measured using clustering accuracy against manually labeled model families as ground truth, representing the proportion of models correctly grouped with their true parent models. \autoref{fig:all_unsupervised_cluster} illustrates the visual results across different strategies, with subfigures showing standard K-Means (\autoref{fig:uncentered-kmeans}), GMM (\autoref{fig:gmm}), Hierarchical Clustering (\autoref{fig:hierarchical}), and DBSCAN (\autoref{fig:dbscan}) projections.

\begin{table}[ht]
\centering
\caption{Accuracy comparison of clustering strategies.}
\label{tab:clustering-comparison}
\begin{tabular}{r|c}
\toprule
\textbf{Clustering Strategy} & \textbf{Accuracy} \\
\midrule
\textbf{Centroid-Initialized K-Means} & \textbf{94.0\%}\\
Hierarchical Clustering & 78.0\%\\
GMM & 76.0\%\\
Standard K-Means& 54.0\%\\
DBSCAN & 44\%\\
\bottomrule
\end{tabular}
\end{table}

As demonstrated in \autoref{tab:clustering-comparison}, centroid-initialized K-Means achieves 94\% accuracy (see \autoref{fig:cluster_result}), followed by hierarchical clustering at 78\%, GMM at 76\%, and DBSCAN at just 44\%. The superior performance of centroid-initialized clustering stems from its ability to leverage prior knowledge of base model fingerprints as initialization points, eliminating the ambiguity inherent in random or data-driven initialization strategies. Traditional clustering methods struggle with overlapping embeddings and boundary cases, particularly when fine-tuned variants exhibit subtle fingerprint variations that confound automatic cluster assignment.

The substantial performance gap between centroid-guided and conventional clustering approaches validates our design decision. When base model fingerprints are available, incorporating them as initialized reference points provides significantly more reliable model attribution than data-driven clustering initialization. This finding is particularly critical for applications requiring high precision, such as license enforcement and provenance verification.

\find{
\textbf{Answer to RQ3:} Experimental results confirm that centroid-initialized K-Means clustering strategies are substantially more effective than conventional clustering alternatives, achieving 94\% accuracy compared to 82\% for the best conventional method.
}

\section{Discusstions}
\label{sec:limitations}

\subsection{Implications}
Through our case analyses, we observe that even minor fingerprinting under limited perturbation is sufficient for accurate attribution, suggesting that LLMs possess stable and distinguishable gradient-level characteristics. This insight highlights a shift from conventional weight-centric comparison to behavioral feature profiling, offering a more interpretable approach to model similarity detection. \tool{} bridges traditional parameter comparison and output-based behavior probing by tracing how input perturbations propagate through model gradients, capturing structural modifications like LoRA integration more effectively than output-only methods while remaining robust to parameter transformations such as quantization or reordering, unlike raw tensor matching techniques that require identical model formats. The gradient-based fingerprinting thus provides a unified framework that combines the structural awareness of parameter-level methods with the flexibility of behavior-based approaches, enabling accurate similarity assessment without requiring private training data.

\subsection{Limitations}
\noindent\underline{\textbf{Limited model scope and efficiency trade-off.}}
Our current evaluation focuses on fine-tuned LLMs of manageable size ($\leq$13B) due to hardware limitations. Processing a single model requires over 20 GB of GPU memory and takes one hour, which significantly limits scalability compared to simple hash-based methods or embedding comparison. These computational constraints restricted our experimental scope\footnote{Similar works~\cite{zhang2024reef,zeng2024huref} evaluate their methods on at most 51 models.} and prevent real-time or large-scale scanning applications unless optimization strategies are introduced. Larger-scale foundation models, multi-modal architectures, or hybrid graph-weighted variants remain unexplored. 

\noindent\underline{\textbf{Gradient sensitivity versus semantic similarity.}}
A noteworthy phenomenon is that models with extremely similar downstream behavior may still diverge in their gradient-level fingerprints due to subtle architectural changes. While this enhances our ability to detect tampering or fine-tuning, it also poses challenges for grouping models with divergent parameter layouts.

\noindent\underline{\textbf{Difficulty in distinguishing structurally similar models.}}
During evaluation, we observed that certain models, such as \textit{Gemma-2B}~\cite{gemma2_2b_it} and \textit{LLaMA-3.1-3B}, produce remarkably similar fingerprint representations, despite being released by different organizations and trained with separate datasets. \textbf{This suggests that models sharing similar transformer backbone configurations and pretraining strategies may converge to comparable gradient response patterns under perturbation}.
This phenomenon indicates a fundamental limitation of fingerprint-based detection: \textbf{when models are architecturally indistinct, gradient-based fingerprints alone may provide insufficient discriminative power for reliable attribution}. While our method remains effective in detecting explicit fine-tuning or aggressive merging, it may struggle in attribution scenarios involving independently trained but structurally similar models. This limitation underscores the need for enhanced feature representations that incorporate additional context--such as training corpus metadata, tokenizer alignment, or attention map dynamics--to further disambiguate seemingly similar model variants.

\subsection{Future Directions}

There are several promising directions to extend this work. First, enhancing fingerprint robustness through methods invariant to minor perturbations and aware of low-rank transformations would enable more reliable detection of fine-tuned or obfuscated models. Second, accelerating fingerprint extraction via scalable inference techniques could significantly reduce computational overhead without compromising accuracy. Third, improving compatibility across diverse model formats (PyTorch, ONNX, GGUF, quantized representations) would facilitate broader industrial adoption. Finally, incorporating sensitivity to deployment-level optimizations such as quantization and operator fusion could enhance model provenance tracing even after aggressive optimization processes.

\section{Related Work}
\label{sec:related work}

Fingerprinting is essential for identifying the origin and lineage of LLMs, especially under unauthorized fine-tuning or parameter merging, which may compromise intellectual property and model integrity. Existing techniques fall into two categories: watermark-based and intrinsic fingerprinting.

\noindent\underline{\textbf{LLM Similarity Detction.}} 
Watermarking techniques embed ownership signals during training, either through trigger-based mechanisms~\cite{xu2024instructional} or prompt-response hashing~\cite{russinovich2024hey}. However, these methods typically require model modification~\cite{li2019piracy}, may degrade performance, and remain susceptible to post hoc alterations such as fine-tuning. Thus, they are unsuitable for post-release attribution.In contrast, intrinsic approaches analyze model outputs~\cite{yang2024fingerprint,mcgovern2024your} or internal representations~\cite{zhang2024reef,zeng2024huref} without altering training. While non-intrusive, they often assume access to internal layers or aligned inputs—assumptions that may not hold for publicly shared models.

\noindent\underline{\textbf{LLM Family Classification.}} Understanding the lineage of LLMs is vital for auditing and attribution. Horwitz et al.~\cite{horwitz2024origin} introduce the \textit{Model Tree} to formalize LLM family classification, proposing the MoTHer task to recover fine-tuning hierarchies from model weights. Their method estimates \textit{node distances} via weight differences and infers \textit{edge directions} using kurtosis-based monotonic trends over training. MoTHer effectively identifies related models, even under parameter-efficient tuning such as LoRA~\cite{hu2022lora}, by leveraging low-rank-aware metrics. Validated on curated benchmarks and real-world LLMs like Llama 2, this work offers a principled approach to tracing model provenance via weight-space analysis.

\section{Conclusion}
\label{sec:conclusion}

This paper addresses the critical gap in LLM provenance tracking by introducing \tool{}, a gradient-based fingerprinting framework for model similarity detection. Our approach extracts behavioral signatures through gradient analysis, operating independently of training data or watermarks. Evaluation on 58 models across five families demonstrates 94\% classification accuracy, providing a foundation for license compliance verification and unauthorized derivation detection in modern LLM ecosystems.

\bibliographystyle{IEEEtran}
\bibliography{main}

\end{document}